\documentclass[10pt,twocolumn,letterpaper]{article}

\usepackage[pagenumbers]{cvpr} % To force page numbers, e.g. for an arXiv version

\usepackage[dvipsnames]{xcolor}

\usepackage{nicematrix}
\usepackage{multirow}  % Include this package

\definecolor{cvprblue}{rgb}{0.21,0.49,0.74}
\usepackage[pagebackref,breaklinks,colorlinks,citecolor=cvprblue]{hyperref}

\def\paperID{12334} % *** Enter the Paper ID here
\def\confName{CVPR}
\def\confYear{2025}

\title{Semantic Segmentation Prior for Diffusion-Based  Real-World Super-Resolution}
\author{Jiahua Xiao$^1$, Jiawei Zhang$^2$, Dongqing Zou$^2$, Xiaodan Zhang$^3$, Jimmy Ren$^2$, and Xing Wei$^{1}$\\
{\small $^1$Xi'an Jiaotong University, $^2$SenseTime Research, $^3$Beijing University of Technology}\\
}

\begin{document}
\maketitle
\begin{abstract}

Real-world image super-resolution (Real-ISR) has achieved a remarkable leap by leveraging large-scale text-to-image models, enabling realistic image restoration from given recognition textual prompts.
However, these methods sometimes fail to recognize some salient objects, resulting in inaccurate semantic restoration in these regions.
%
% Additionally, these methods rely on the implicit cross-attention mechanism for semantic prompt localization, which may  overallocate attention to irrelevant prompts within the same spatial region, leading to semantic ambiguity in image generation.
%
Additionally, the same region may have a strong response to more than one prompt and it will lead to semantic ambiguity for image super-resolution. 
%
% Despite the success in real-world image super-resolution (Real-ISR) with the large-scale pretrained diffusion models, existing approaches still struggle to achieve region-precise semantic perception and control. 
% For example, 
% existing text-prompt guided methods fail to achieve effective semantic spatial localization relying on cross-attention mechanism, leading to semantic regions perception errors or semantic mixing. 
% Additionally, the fine-grained semantic layout text descriptions and understanding such complex text descriptions for models are challenging to achieve. Insights from semantic segmentation, which integrates pixel-level semantic class labels with fine-grained spatial layout,
% For example, 
% existing text-prompt guided methods fail to achieve effective semantic spatial localization relying on an implicit cross-attention mechanism, leading to semantic region perception errors. Additionally,  text prompts based on multimodal language models often struggle to capture all objects within an image, causing less prominent objects to be missed or underemphasized, 
% leading to uneven detail enhancement across the image.
%
To alleviate the above two issues, in this paper, we propose to consider semantic segmentation as an additional control condition into diffusion-based image super-resolution.  Compared to textual prompt conditions, semantic segmentation enables a more comprehensive perception of salient objects within an image by assigning class labels to each pixel. It also mitigates the risks of semantic ambiguities by explicitly allocating objects to their respective spatial regions.
In practice, inspired by the fact that image super-resolution and segmentation can benefit each other, we propose SegSR which introduces a dual-diffusion framework to facilitate interaction between the image super-resolution and segmentation diffusion models. 
Specifically, we develop a Dual-Modality Bridge module to enable updated information flow between these two diffusion models, achieving mutual benefit during the reverse diffusion process.
% to acquire segmentation conditions for the Real-ISR diffusion models, we train a degradation-robust diffusion segmentor for semantic segmentation prior modeling, which strives to produce reasonable segmentation predictions.
%
% Finally, SegSR facilitates dynamic interaction between the Real-ISR and semantic segmentation diffusion models at each iterative step. 
%
Extensive experiments show that SegSR can generate realistic images while preserving semantic structures more effectively.

\end{abstract}    
\section{Introduction}
\label{sec:intro}
Real-world image super-resolution is a longstanding challenge, as it must handle unknown complex degradations (\eg, low resolution, blur, noise and \etc) while generating perceptually realistic high-quality (HQ) images. 
Classical discriminative approaches~\cite{dong2015image,kim2016accurate,zhang2022efficient,lu2022transformer,chen2023activating} simply assume known degradations and often produce over-smoothed results.
% , making them unsuitable for real-world scenarios.
%
Despite significant achievements in improving visual perception, generative adversarial network (GAN)-based methods~\cite{bsrgan,realesrgan,liang2022efficient,park2023content} struggle to balance perceptual quality with fidelity perversion and often result in artifacts or distorted details due to the instability of adversarial training. 

Recently, the advent of diffusion models (DMs)~\cite{ddpm} has demonstrated impressive capabilities in approximating complex distributions and generating realistic images\cite{dhariwal2021diffusion}. 
%
% As a result, DMs are gradually emerging as promising alternatives to GANs~\cite{gan}.
%
Especially with the advent of large-scale pretrained text-to-image (T2I) models like StableDiffusion (SD)~\cite{sd}, image generation has advanced to a new stage of development.
%
% In this context, researchers have increasingly focused on leveraging the powerful SD generative priors to address the challenges of Real-ISR problem.
%
In this context, researchers have increasingly focused on leveraging the powerful generative ability of StableDiffusion to improve the restoration performance of SR.
%
% Among these diffusion-based super-resolution methods,  StableSR~\cite{stablesr} applies generative priors for image super-resolution by utilizing the latent representation of LQ images as the control condition.
%
Among these diffusion-based super-resolution methods,  StableSR~\cite{stablesr} utilizes the latent representation of LQ images as the control condition to guide StableDiffusion for super-resolution.
In contrast, DiffBIR~\cite{diffbir} first performs an initial restoration of LQ images before leveraging generative priors to balance the quality and fidelity in the diffusion process.
These methods highlight the substantial potential of generative priors in SR tasks, 
yet using solely LQ image information without additional semantic control may lead to incorrect content reconstruction.

Given the inherent advantages of textual prompts in guiding generation within pre-trained T2I models, recent methods~\cite{xpsr,spire,adadiffsr,supir} have shifted towards leveraging semantic descriptions derived from LQ images to further control the generation process in StableDiffusion and improve the semantic fidelity of the image restoration.
For example, PASD~\cite{pasd} and SeeSR~\cite{seesr} utilize pretrained captioning~\cite{li2023blip} or tagging~\cite{ram} model to extract image content, where the textual semantic descriptions of image content serving as prompt conditions to guide the generation process.
However, these methods might exhibit two potential limitations.
Firstly, they sometimes fail to recognize some salient components and generate inaccurate semantic details in these regions.
%
% Secondly, the implicit cross-attention mechanism for semantic prompt localization can overallocate attention to irrelevant prompts, leading to semantic ambiguity in image generation.
Secondly, the same region may have a strong response to more than one prompt which will lead to semantic ambiguity for image super-resolution.
(Please refer to Section~\ref{sec:motivation} for more discussions.)

In this paper, we aim to alleviate the above two issues by leveraging semantic segmentation as an additional control condition.
%
% Compared to textual prompts, semantic segmentation can not only assign a unique class label to each pixel, explicitly aligns semantic information with specific spatial regions and assigns a unique class label to each pixel, facilitating clearer semantic spatial localization while mitigating the risks of semantic ambiguities and object omissions.
Compared to textual prompt conditions,  semantic segmentation enables a more comprehensive perception of salient objects within an image by assigning class labels to each pixel.
It also facilitates clearer semantic spatial localization and mitigates the risks of semantic ambiguities by explicitly allocating objects to their respective spatial regions.
However, in practice, predicting segmentation masks from severely degraded images is challenging, and inaccurate segmentation masks can lead to incorrect semantic content and distorted spatial structures in the restoration result. 

While directly finetuning segmentation network (e.g., SegFormer~\cite{segformer}, SegNext~\cite{segnext}) on LQ images can acquire degradation-aware capabilities to some extent, the performance is still limited.
%
% it is essential to highlight that the iteratively improved image results can enhance segmentation predictions during the diffusion process, and vice versa. 
%
% Inspired by this property,  we propose SegSR (see Figure~\ref{fig:framework}), an effective semantic region perception method, which introduces a dual-diffusion framework to facilitate interaction between the Real-ISR and segmentation diffusion models for optimal cooperative restoration.
%
Motivated by the fact that image super-resolution and segmentation can benefit each other, we propose SegSR (see Figure~\ref{fig:framework}) which introduces a dual-diffusion framework to facilitate interaction between the image super-resolution and segmentation diffusion models.
Specifically, it consists of three parts: (i) the super-resolution diffusion (SRDM) branch, based on StableDiffusion, generates realistic restoration images conditioned on the LQ as well as segmentation information.
%
% (ii)  The segmentation diffusion (SegDM) branch models the ground truth mask prior using a discrete diffusion model~\cite{ddm}. It iteratively refines the quality of incorporated degradation-aware segmentor predictions, ensuring better alignment with the ground truth mask prior distribution,  as discussed in~\cite{ddps}.
(ii) The segmentation diffusion (SegDM) branch, based on discrete diffusion model~\cite{ddm}, to estimate a more accurate segmentation mask with the guide from image information.
%
% (iii) The Dual-Modality Bridge (DMB) enables the information flow between the SRDM and SegDM branches, incorporating the intermediate image restoration and segmentation refinement outcomes into each other's diffusion processes in all the diffusion steps.
%
(iii) The Dual-Modality Bridge (DMB) enables the information flow between the SRDM and SegDM branches so that these two branches can benefit each other during the reverse diffusion process.

Our contributions can be summarized as follows

\begin{itemize}
    % \item We explore the segmentation semantic priors to alleviate the limitations of salient object recognition omissions and semantic ambiguity in image generation, as exposed in existing prompt-based Real-ISR methods. 
    \item We explore the segmentation semantic priors to guide diffusion-based image super-resolution generation. 
    % \item To effectively integrate the structured semantic prior, we embed them within a Seg2Img generator to ensures precise alignment between semantic label and spatial layout during content generation.
    \item We propose a collaborative framework, termed as SegSR that facilitates mutual cooperation between SRDM and SegDM at each reverse diffusion step. 
    \item Extensive experiments demonstrate that SegSR produces realistic images while effectively preserving semantic structures.

\end{itemize}

\section{Related Work}
\label{sec:related}
\begin{figure*}[!t]
    \centering
    \includegraphics[width=\linewidth]{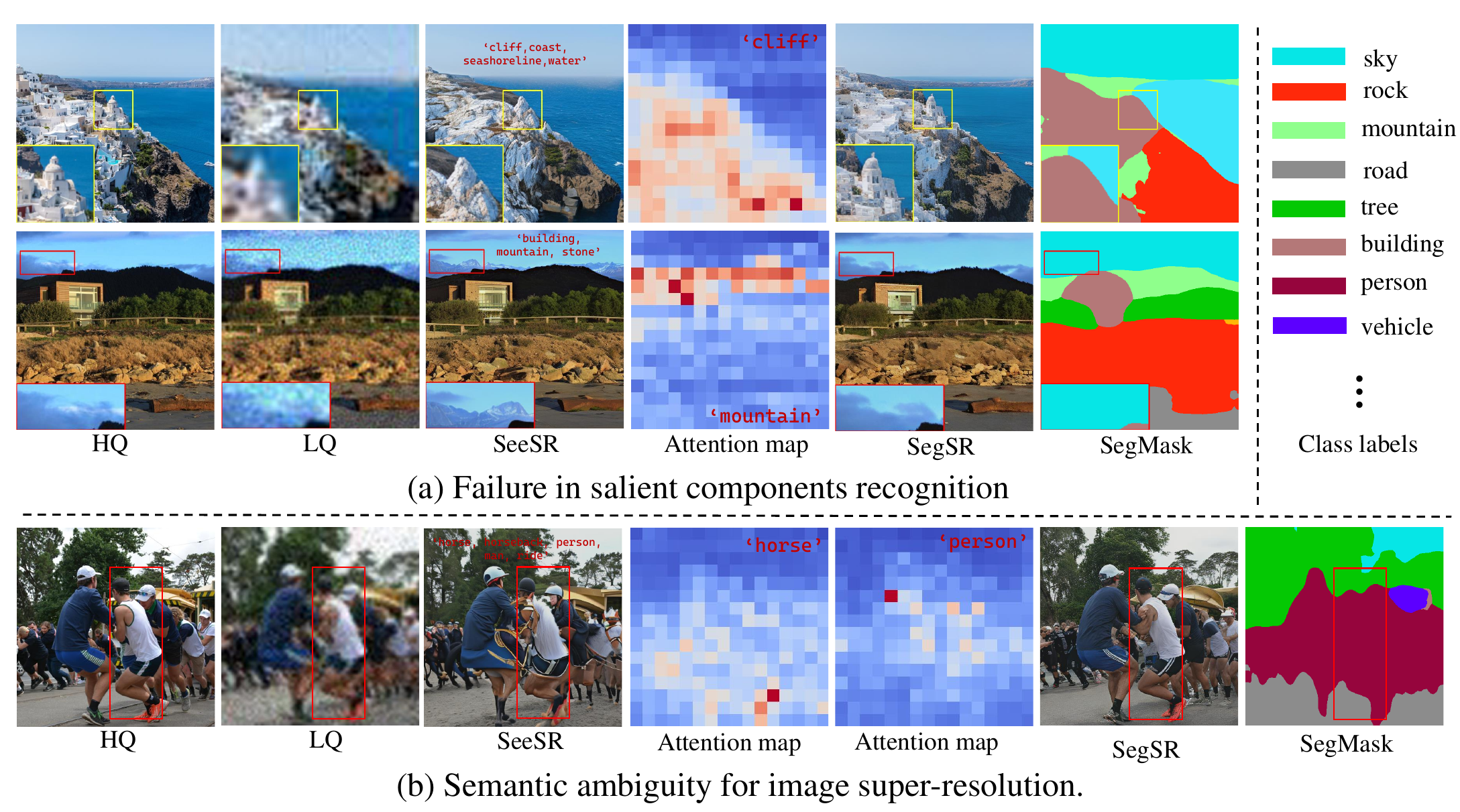}
    \caption{Comparison of Real-ISR results between SegSR conditioned on segmentation masks and prompt-guided methods (examplifed by SeeSR~\cite{seesr}). 
   (a) SeeSR fails to recognize some salient components in the image. The cross-attention maps show the attention weight allocation of other objects in the region, leading to the inaccurate generation of semantic details.
   % displays the cross-attention map of irrelevant object prompts in the zoomed region when corresponding salient object prompts are missing, showing incorrect attention allocation in the region that leads to semantically inaccurate image generation.
    %
    % In contrast, SegSR uses semantic segmentation as an additional control condition, enabling a more comprehensive object perception and generates semantically accurate content within corresponding regions.
    %
    (b) The same region have a strong response to more than one prompt  through cross-attention and it lead to semantic ambiguity outcomes for image restoration. 
    In the cross-attention map visualization, \textcolor{red}{warmer} color indicate higher attention weights, while \textcolor{blue}{cooler} color represent lower attention weights.
     In contrast, SegSR can restore more faithful details as long as the estimated segmentation mask is accurate.
     % SegSR explicitly allocating separate spatial regions for each object, ensuring clearer semantic understanding.
     }
    \label{fig:errors}
\end{figure*}
{\flushleft\textbf{Diffusion Probabilistic Models.}}\quad
With the emergence of DDPM~\cite{ddpm} and DDIM~\cite{ddim}, Diffusion models have become the new benchmark for tasks such as image synthesis~\cite{dhariwal2021diffusion} and editing~\cite{kawar2023imagic}. Compared to GANs~\cite{gan}, diffusion models offer more stable training and superior generation quality, marking a significant advancement in image generation.
Due to the significant efficiency issues arising from the same-sized latent code as the original image in DDPM, the advent of StableDiffusion (SD)~\cite{sd} has further enhanced efficiency by operating in a compressed latent space, significantly reducing computational costs while maintaining high quality. 
For example, StableDiffusion-v2.1\footnote{\url{ https://huggingface.co/stabilityai/stable-diffusion-2-1}} 
is trained on extensive datasets comprising over 5 billion image-text pairs, endowing it with robust natural image priors.
% Moreover, to control the adherence of diffusion models to prompts during the sampling process, researchers have introduced classifier-free guidance (CFG)~\cite{cfg} to enhance the flexibility of controlled generation.

{\flushleft\textbf{Generative Super-Resolution.}}\quad
% Real-ISR aims to restore complex and unknown degraded low-resolution images in real-world scenarios. 
Given the challenges in obtaining real-world LQ-HQ image pairs, recent advancements (e.g., BSRGAN~\cite{bsrgan}, Real-ESRGAN~\cite{realesrgan}) attempt to design degradation pipelines to simulate real-world degradation and utilize GANs to restore realistic HR images. However, the inherent instability and mode collapse in GAN-based methods has shifted increasing focus toward diffusion models for more reliable and diverse image restoration. 
Early attempts~\cite{sr3,srdiff,resshift} to train diffusion models from scratch in the pixel domain conditioned on LQ images are not only resource-consuming but also failed to leverage the generative priors of existing pre-trained diffusion models. Recently, researchers~\cite{stablesr,diffbir,pasd,seesr,supir,xpsr} have begun utilizing powerful pre-trained T2I models to tackle the challenge of Real-ISR. Among them, StableSR~\cite{stablesr} conditions on LQ images and uses a time-aware encoder to guide the generation process in StableDiffusion. DiffBIR first performs an initial restoration for LQ images, then trains a ControlNet~\cite{controlnet} for further refinement.

Rather than relying solely on LQ images as the control condition for StableDiffusion, more recent approaches have focused on extracting semantic information from images and incorporating semantic textual prompts through cross-attention as an additional condition. For example, 
PASD~\cite{pasd} leverages  BLIP~\cite{blip} to obtain image caption information from the LQ image. SeeSR~\cite{seesr} employs a degradation-aware RAM~\cite{ram} to extract more concise tag-style prompts. SUPIR~\cite{supir} utilizes the multi-modal large language model (i.e., LLaVA~\cite{llava}) to provide precise image content prompts to improve the semantic accuracy of restored images. However, these
prompt-guided Real-ISR methods sometimes fail to recognize some salient components and struggle with clear semantic spatial localization.

{\flushleft\textbf{Semantic Segmentation Prior.}}\quad
Semantic segmentation can be viewed as a pixel-level classification task.  
classic discriminative works~\cite{segformer,segnext} adopt an encoder-decoder architecture to assign a semantic label to each region in an image. In contrast to the class-agnostic segmentation approach of SAM, open vocabulary segmentation methods, such as X-decoder~\cite{x-decoder} and SEEM~\cite{seem}, leverage user-provided text inputs to generate masks for segmentation tasks. Most recently, DDPS~\cite{ddps} enhances semantic segmentation quality with a mask prior modeled by a discrete diffusion model~\cite{ddm}. As semantic segmentation provides critical semantic and spatial information, in the past, some works have used segmentation masks as a prior to guiding more accurate image restoration. For instance, SFTGAN~\cite{sftgan} demonstrates the effectiveness of segmentation prior to recovering textures faithful to semantic classes.   SSG-RWSR~\cite{ssg-rwsr} proposes to guide the learning of the super-resolution learning process with the loss of a semantic segmentation network. SAM-DiffSR~\cite{sam-dffsr} improves different image areas by modulating the diffusion noise distribution by class-agnostic segmentation masks generated by SAM~\cite{sam}. However, achieving accurate semantic segmentation on real-world LQ images with unknown severe degradation, and effectively utilizing such segmentation priors to enhance the performance of Real-ISR tasks, remains a significant challenge.

\section{Methodology}
\label{sec:method}
\begin{figure*}[!t]
    \centering
    \includegraphics[width=\linewidth]{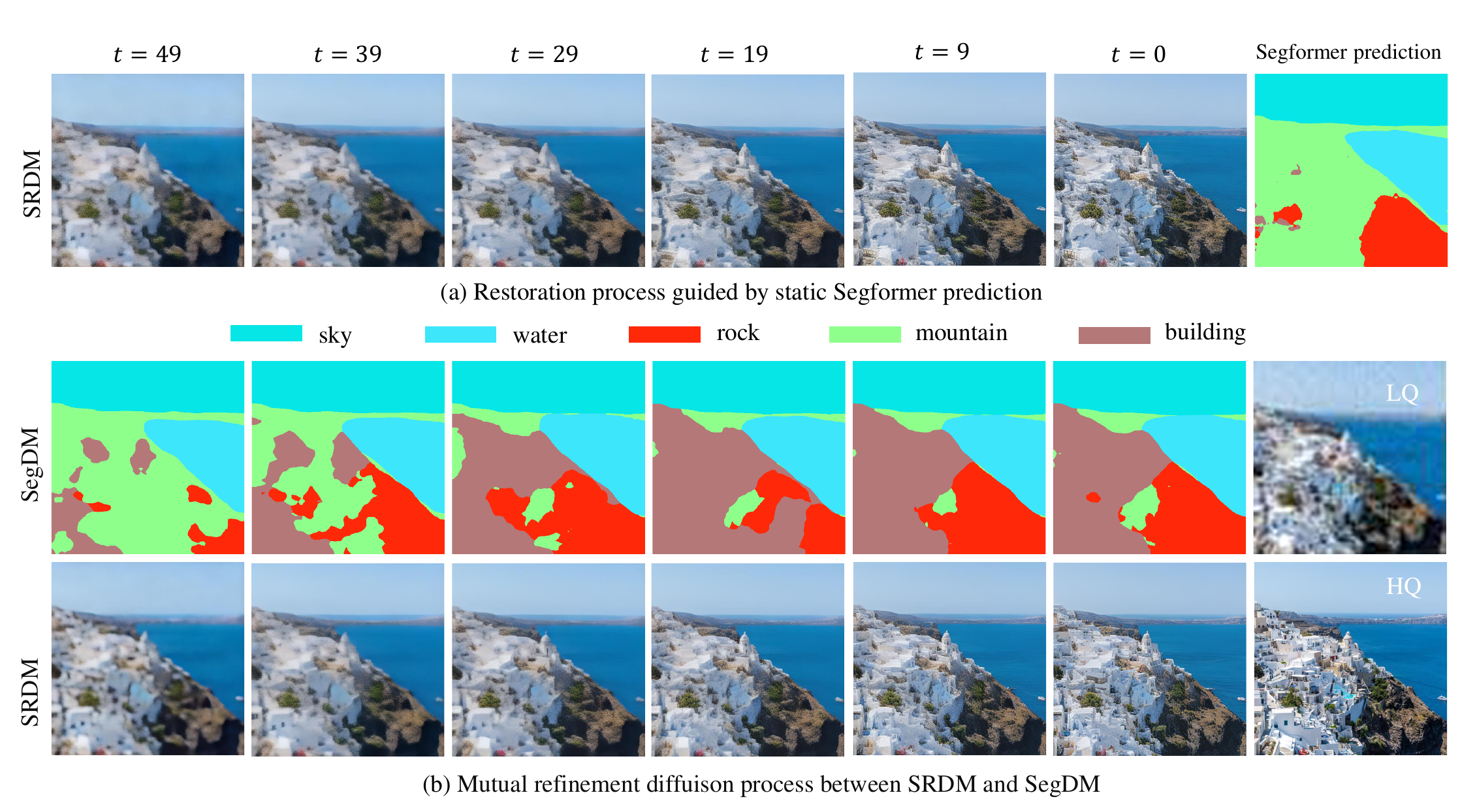}
    \caption{\textbf{Mutual Refinement within SegSR}. We present the final result predictions at different steps $t$ of the inverse diffusion process for both SRDM and SegDM. (a) To provide semantic segmentation priors for Real-ISR, the pretrained Segformer~\cite{segformer} predicts segmentation masks from degraded images, but these predictions become inaccurate when the degradation is severe. As a result, the Segformer-guided SRDM struggles to restore images with high semantic fidelity. (b) The proposed SRDM and SegDM mutually benefit from each other through the DMB in SegSR, progressively improving segmentation predictions and image quality through the inverse diffusion process.}
    \label{fig:inter}
\end{figure*}
\subsection{Motivation}
\label{sec:motivation}
To harness the potential of pretrained T2I diffusion models, methods like PASD~\cite{pasd} and SeeSR~\cite{seesr} successfully integrate high-level semantic prompt information as additional conditions to restore more realistic images.
%
% These methods enable Real-ISR to generate images with more realistic details and improved semantic fidelity.
%
However, these prompt-based image super-resolution methods rely on multimodal language models to first recognize the semantic contents of images and then implicitly localize semantic prompts through cross-attention.
This will have two potential limitations.
% Firstly, these prompts-based methods lack detailed spatial position descriptions,  which can result in misalignment between prompts and regions in the LQ image.
%
% Firstly, multimodal language models often overlook to describe some salient objects in the image, leading to inaccurate restoration of semantic details in regions corresponding to those omited objects.
First of all, multimodal language models sometimes fail to recognize some salient components in the image.
As a result, these salient components will be restored with inaccurate semantic details.
As shown in Figure~\ref{fig:errors}~(a), SeeSR fails to recognize the building as well as the sky and the cross-attention mistakenly considers them as `cliff' and `mountain' which will lead to cliff-like and mountain-like textures in the building and sky regions in the restoration results respectively.
%
% localizes the object `stone' to the building-styled region, causing stone-like textures to appear in the building area.
%
% In addition, as shown in the bottom row of Figure~\ref{fig:errors} (a), SeeSR fails to capture salient the object `sky' or `cloud' across the image. This leads to inaccurate restoration of the zoomed area,  as cross-attention misallocates attention from the object 'mountain' to this region.
%
% The other issue is that the implicit localization through cross-attention may overallocate weights to more than one token within the same spatial region which will lead to semantic ambiguity in image restoration.
The other issue is that the same region may have a strong response to more than one token through cross-attention and it will lead to semantic ambiguity for image restoration.
%
% As shown in Figure~\ref{fig:errors}~(b), cross-attention assigns weights to both `horse' and `person' in the red-boxed area, resulting in a confusing image that blends `person' and `horse' features in this region. 
%
As shown in Figure~\ref{fig:errors}~(b), the red-box region has a strong response for both `horse' and `person' and the restoration result from SeeSR has a blending of `person' and `horse' style in that region.
% In addition, the implicit localization of cross-attention may lead to overallocate weights to irrelevant tokens within a same spatial region, confusing the model's understanding of this area and leading to semantic ambiguous generation. As shown in Figure~\ref{fig:errors}(b), cross-attention assigns weights to both "horse" and "person" in the red-boxed area, resulting in a confusing image that blends "person" and "horse" features in this region. For the second issue, multimodal language models often prioritizes more salient objects, resulting in incomplete descriptions in high-density object scenes. Consequently, the omitted object regions lose semantic control, and cross-attention may even allocate weights from other objects to these regions. As shown in  bottom row of Figure~\ref{fig:errors}(a), The DAPE recognition module in SeeSR fails to cover the "sky" class prompt, resulting in the zoomed area not being accurately restored and misallocate the prompt of "mountain" to this region.
%

To alleviate the above issues raised by implicit localization through cross-attention, we propose to consider semantic segmentation into diffusion-based image super-resolution.
For semantic segmentation, each pixel will be explicitly assigned to a label so that almost all the salient components from the input will be recognized without omission. 
Specifically, the proposed SegSR considers information from semantic segmentation as an additional condition to guide the reverse diffusion process for realistic image super-resolution.
As shown in Figure~\ref{fig:errors}, the proposed SegSR can restore more faithful details as long as the estimated segmentation mask is accurate.

% This motivates us that if we can introduce semantic segmentation into Real-ISR, which holds the potential to enable more comprehensive perception of salient objects within the image by assigning class labels to each pixel. 
% %
% Additionally, it facilitates clearer semantic spatial localization and reduces the risks of semantic ambiguity by explicitly allocating objects to their respective spatial regions.
% % %
% % The semantic segmentation prior holds the potential to enable more comprehensive perception of salient objects within the image, while facilitating clearer semantic spatial localization and reducing the risks of semantic ambiguities.
% %
% % As shown in Figure~\ref{fig:errors}, our method incorporated with semantic segmentation priors can effectively address the aforementioned limitations of prompt-guided methods, generating the correct and coherent semantic content in each corresponding area. 
% %
% As shown in Figure~\ref{fig:errors}, our method, incorporating semantic segmentation priors, effectively addresses the aforementioned limitations of prompt-guided methods and generates accurate, coherent semantic content in each corresponding area.

\begin{figure*}[!t]
    \centering
    \includegraphics[width=\linewidth]{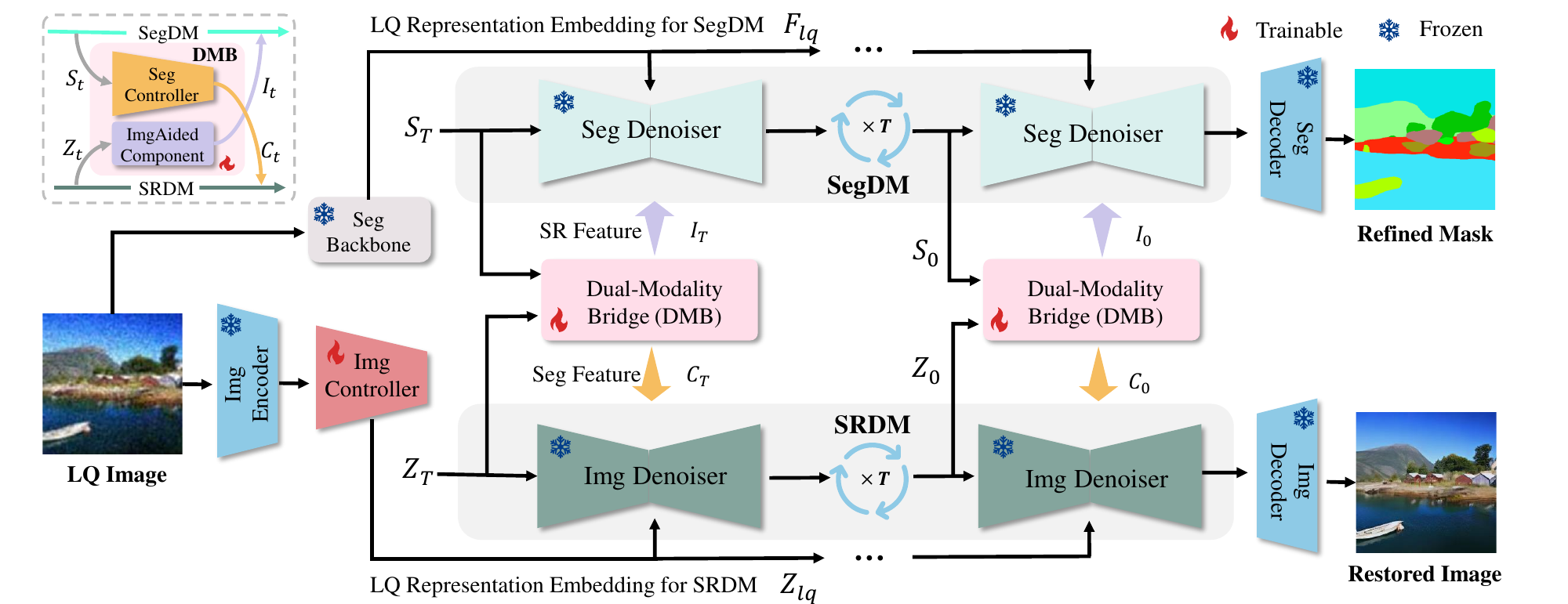}
    \caption{\textbf{Overview of SegSR}.  Framework comprises three key parts: i) SRDM performs super-resolution diffusion process, conditioned on LQ image embeding $Z_{lq}$ and gradually updated segmentation prior $S_t$ from SegDM to generate high-realness image; ii) SegDM conducts semantic segmentation diffusion process, conditioned on LQ image features $F_{lq}$ and iteratively restored image information $Z_t$ from SRDM to improve the accuracy of segmentation priors; (iii) the DMB module, which encodes intermediate updated features $Z_{t-1}$ and $S_{t-1}$ from SRDM and SegDM from the previous step, producing the image and segmentation conditions $I_{t}$ and $C_{t}$ for the current time step. SRDM and SegDM collaborate through the DMB module to ultimately achieve realistic image super-resolution. }
    \label{fig:framework}
\end{figure*}

However, directly estimating the segmentation mask from the LQ image is challenging.
As shown in Figure~\ref{fig:inter}~(a), the segmentation mask estimated from Segformer is inaccurate even though it is finetuned based on LQ images.
As a result, the restoration result guided by this inaccurate mask will contain unrealistic details (\eg, generating `mountain' textures in the building).
%
% As shown in Figure~\ref{fig:inter}~(a), 
% we use a Segformer pretrained on LQ images to predict segmentation masks that guide diffusion-based ISR model (SRDM) in restoration.
% %
% However, its predictions on degraded images are often inaccurate, leading SRDM to generate incorrect semantic content in corresponding regions (e.g., generating `mountain' textures in `building' regions). 
%
Motivated by the fact that image semantic segmentation and diffusion-based image super-resolution can benefit each other, we propose SegSR which is a dual-branch framework that contains two diffusion models, which are denoted as diffusion-based super-resolution (SRDM) and diffusion-based segmentation (SegDM), for these two tasks respectively.
Simplistically speaking, SRDM can utilize the updated segmentation information from SegDM and vice versa during the reverse diffusion process through a proposed Dual-Modality Bridge (DMB) which is shown in Figure~\ref{fig:framework}.
% It is worth emphasizing that improved image results and refined segmentation predictions can mutually enhance each other's quality.
%
% This motivates us to leverage restored features from SRDM to assist semantic segmentation, while refined segmentation predictions from diffusion-based segmentor (SegDM) to enhance image restoration.
% %
% Therefore, in this work, we attempt to design a mutual refinement framework that facilitates interaction in both SRDM and SegDM, ultimately achieving image restoration performance.
%
% As shown in Figure~\ref{fig:inter} (b), the dual-diffusion mutual refinement framework can gradually restore HQ image and more accurate semantic segmentation predictions with the information interaction between the SRDM and SegDM across all steps.
As shown in Figure~\ref{fig:inter}~(b), both the segmentation and SR can get better results during the reverse diffusion process.
In the following subsections, we will describe the details of the proposed SegSR based on the above two motivations.

\subsection{Framework Overview}
\label{sec:overview}
In this paper, we present SegSR, a semantic segmentation prior-interpolated diffusion framework designed to tackle the challenge of image super-resolution. As shown in Figure~\ref{fig:framework}, SegSR consists of two branches, SRDM and SegDM, connected by the Dual-Modality Bridge (DMB). Among them, SRDM employs the architecture of StableSR to generate realistic restored images based on the conditions of the LQ image. SegDM is a discrete diffusion model that gradually predicts the segmentation mask from the given LQ image. As for DMB, it acts as a bridge connecting SRDM and SegDM, based on the fact that image super-resolution and segmentation can benefit each other. During the diffusion process, it provides updated segmentation priors to SRDM and iteratively restored image information to SegDM. For more detailed information about these modules, please refer to the supplementary material.

\subsection{Dual-Diffusion models with Mutual Refinement}
% As discussed in the subsection~\ref{sec:overview},   
% predicting segmentation masks from low-quality images is challenging, and the restoration result guided by this inaccurate mask will contain unrealistic details.
% %
As discussed in Section~\ref{sec:motivation}, image super-resolution and segmentation can benefit each other. In this subsection, we present a dual-diffusion framework (SegSR) to facilitate interaction between diffusion-based 
super-resolution (SRDM) and diffusion-based segmentation
(SegDM) at each diffusion step.
It consists of two branches, SRDM and SegDM, connected by the Dual-Modality Bridge (DMB) as described in Section~\ref{sec:overview}. 

Among them, SRDM performs the diffusion forward and reverse process in the latent space through a pretrained  VAE~\cite{sd} (Imd Encoder and Image Decoder).
After encoding HQ image into the latent embedding $Z_0$, the diffusion process sequentially adds noise into $Z_0$ at step $t$, generating the noisy latent $Z_t$. Then, a noise prediction network~\cite{sd} (Image Denoiser) is used to progressively remove the noise during the reverse process.
To ensure consistency between the restoration results and the input image, the LQ image is first encoded into the latent space using VAE. Then, the LQ latent is fed into the Img controller to obtain the LQ representation embedding $Z_{lq}$, which is used as a condition for the Img Denoiser.

In the SegDM branch,  similar to DDPS~\cite{ddps}, we adopt the discrete diffusion model to characterize the distribution of ground truth segmentation prior.
Following the Markov chain of the diffusion process, the ground truth segmentation mask is encoded as $S_0$ via a simple resize codec. Then, noise is progressively added over $t$ steps to obtain the noisy state $S_t$.
The segmentation diffusion model (SegDM) employs a U-Net~\cite{unet} (Seg Denoiser) to iteratively denoise the current state $S_t$ to the next state $S_{t-1}$. To guide the diffusion process in generating the semantic segmentation of the input image, the SegDM is conditioned on initial predictions of LQ image representations  $F_{lq}$ from a Segformer model.
As a result, the initial segmentation result can be iteratively refined toward the outcome that better matches the ground truth segmentation mask distribution by SegDM.
\begin{table*}[t]
\centering
\resizebox{\textwidth}{!}{%
\begin{NiceTabular}{c|c|ccc|cccccc}
\toprule
% \multirow{2}{*}{Datasets} & \multirow{2}{*}{Metrics} 
% & \multicolumn{6}{c|}{Methods}& \multicolumn{6}{c|}{Methods} \\ 
% \cline{3-14} &
Datasets &Metrics& BSRGAN~\cite{bsrgan} & Real-ESRGAN~\cite{realesrgan} & DASR~\cite{dasr} & StableSR~\cite{stablesr} & ResShift~\cite{resshift} & PASD~\cite{pasd} & DiffBIR~\cite{diffbir} & SeeSR~\cite{seesr} &SegSR \\ \midrule 
\multirow{8}{*}{DIV2K-Val~\cite{div2k}} 
                          & PSNR ↑ &\textcolor{red}{\textbf{21.47}} &20.96&\textcolor{blue}{\underline{21.24}}&20.72&21.53&20.52&20.89&20.47&20.42 \\  
                          & SSIM ↑  &\textcolor{blue}{\underline{ 0.5144}}&\textcolor{red}{\textbf{0.5201}}&0.5110&0.4903&0.5132&0.4856&0.4916&0.4958&0.4659 \\  
                          & LPIPS ↓  &0.4136 & 0.3870&0.4363&0.3842&0.4162& 0.4178&\textcolor{blue}{\underline{0.3661}}& \textcolor{red}{\textbf{ 0.3503}}&0.3769 \\  
                          & DISTS ↓  & 0.2754&0.2599&0.2918&0.2205&0.2567&0.2299& \textcolor{blue}{\underline{0.1991}}&\textcolor{red}{\textbf{0.1891}}&0.2240 \\  
                          % & NIQE ↓                    & 3.9797&\textcolor{red}{\textbf{3.7756}}&3.9508&4.1730&5.7808&4.1914&\textcolor{blue}{\underline{3.8688}}&4.0357&4.4121 \\  
                          & MUSIQ ↑                   & 63.35&64.80&58.09&68.46&61.88&71.90&71.61&\textcolor{red}{\textbf{72.80}}&\textcolor{blue}{\underline{72.29}} \\  
                          & MANIQA ↑                 &0.3558 &0.4104&0.3159&0.4816&0.3843&0.5343&0.5115&\textcolor{blue}{\underline{0.5735}}&\textcolor{red}{\textbf{0.6006}} \\  
                          & CLIPIQA ↑                  &0.5244&0.5986&0.5541&0.6978&0.5893&0.6655&0.7413&\textcolor{blue}{\underline{0.7444}}&\textcolor{red}{\textbf{0.7723}} \\ \midrule 
\multirow{8}{*}{OST-Val~\cite{sftgan}}   
                          & PSNR ↑ &\textcolor{red}{\textbf{22.48}}&21.79&22.12&21.48&22.62&21.63&21.82&21.58&21.13 \\  
                          & SSIM ↑   &\textcolor{red}{\textbf{0.5191}}&0.5094 &0.5066&0.4791&\textcolor{blue}{\underline{ 0.5187}}&0.4886&0.4888& 0.4968&0.4547 \\  
                          & LPIPS ↓  &0.4088&\textcolor{blue}{\underline{0.3762}} &0.4292& 0.3964&0.4292&0.4231&0.3795&\textcolor{red}{\textbf{0.3573}}& 0.4047\\  
                          & DISTS ↓  &0.2490& 0.2249 &0.2517&0.2134& 0.2534&0.2360&\textcolor{blue}{\underline{0.2007}}&\textcolor{red}{\textbf{0.1991}}& 0.2220\\  
                          % & NIQE ↓   &\textcolor{blue}{\underline{4.0431}}&\textcolor{red}{\textbf{ 3.6358}}&3.8576&4.2999&6.1866&4.0747&4.1709&4.0487& 4.8799 \\  
                          & MUSIQ ↑  &63.12& 69.10 &63.74&67.84&62.46&69.88&71.97&\textcolor{blue}{\underline{72.39}}&\textcolor{red}{\textbf{72.54}}\\  
                          & MANIQA ↑ &0.4003&0.4821 &0.3893& 0.4908&0.4398& 0.4901&0.5331&\textcolor{blue}{\underline{0.5568}}&\textcolor{red}{\textbf{0.6371}} \\  
                          & CLIPIQA ↑&0.5101& 0.6170 &0.5742&0.6746&0.6144&0.5771&\textcolor{blue}{\underline{0.7263}}&0.7025&\textcolor{red}{\textbf{0.7668}} \\ \midrule 
\multirow{8}{*}{RealSR~\cite{realsr}} 
                          & PSNR ↑ &26.37&25.68&\textcolor{red}{\textbf{27.01}}&25.26&\textcolor{blue}{\underline{26.38}}&25.74&24.24&25.14&24.61 \\  
                          & SSIM ↑   &\textcolor{blue}{\underline{0.7651}}&0.7614 &\textcolor{red}{\textbf{0.7708}}& 0.7271&0.7567&0.7294&0.6650&  0.7210& 0.6858 \\  
                          & LPIPS ↓  &\textcolor{red}{\textbf{0.2656}}&\textcolor{blue}{\underline{0.2709}} &0.3134& 0.2912&0.3159&0.3136&0.3469&0.3007& 0.3434\\  
                          & DISTS ↓  &\textcolor{blue}{\underline{0.2124}}& \textcolor{red}{\textbf{0.2060}} &0.2202& 0.2114&0.2433&0.2296&0.2300& 0.2224& 0.2349\\ 
                          % & NIQE ↓                   & 5.6390&5.8027& 6.5438&6.0112&6.8774&5.6757&\textcolor{blue}{\underline{5.4961}}&\textcolor{red}{\textbf{5.3983}}&5.5182 \\  
                          & MUSIQ ↑                 & 63.28&60.36&41.20&63.96&60.21&\textcolor{blue}{\underline{69.09}}&68.34&\textcolor{red}{\textbf{69.82}}&67.80 \\  
                          & MANIQA ↑                 & 0.3772&0.3733&0.2441&0.4074&0.3948&0.5088&0.4847&\textcolor{red}{\textbf{0.5407}}&\textcolor{blue}{\underline{0.5233}} \\  
                          & CLIPIQA ↑                  &0.5117 &0.4491&0.3201&0.5768&0.5492&0.5892&\textcolor{red}{\textbf{0.6961}}&0.6701&\textcolor{blue}{\underline{0.6906}} \\ \midrule 
\multirow{4}{*}{RealLQ250~\cite{dreamclear}} 
% & NIQE ↓                   &4.5371 &4.1293&4.7858&3.9376&5.9949&4.2954&4.5404& 4.3849&4.1765 \\  
                          & MUSIQ ↑                 &63.51 &62.51&53.02&68.56&61.18&\textcolor{red}{\textbf{70.73}}&66.07&69.44& \textcolor{blue}{\underline{69.48}} \\  
                          & MANIQA ↑                 &0.3479 &0.3543&0.2720& 0.4658&0.3761&\textcolor{blue}{\underline{0.4805}}&0.4269& 0.4742&\textcolor{red}{\textbf{0.5040}} \\  
                          & CLIPIQA ↑                  & 0.5691&0.5435&0.4631&0.7267&0.6237&\textcolor{blue}{\underline{0.6932}}&0.6821&0.6796&\textcolor{red}{\textbf{0.7381}} \\ \bottomrule 
\end{NiceTabular}%
}
\caption{Quantitative comparison of SOTA methods across four datasets. The best and second-best results are marked in \textbf{\textcolor{red}{bold}} and \textcolor{blue}{\underline{underline}}.}
\label{tab:quantitative}
\end{table*}

\def\fwidth{0.1215\linewidth}
\def\arraystretch{0.5}
\renewcommand{\tabcolsep}{0.5 pt}
\begin{figure*}[!ht]
\centering
\begin{tabular}{ccccccccc}
    % \centering{\rotatebox{90}{  \small{DIV2K-810}}}  \hspace{1pt} &
    % \includegraphics[width=\fwidth]{figs/results/div2k-val/810_color_rect/LQ.png} &  
    % \includegraphics[width=\fwidth]{figs/results/div2k-val/810_color_rect/bsrgan.png} &  
    % \includegraphics[width=\fwidth]{figs/results/div2k-val/810_color_rect/realesrgan.png} &  
    % % \includegraphics[width=\fwidth]{figs/results/div2k-val/810_color_rect/dasr.png} & 
    % \includegraphics[width=\fwidth]{figs/results/div2k-val/810_color_rect/stablesr.png} &  
    % \includegraphics[width=\fwidth]{figs/results/div2k-val/810_color_rect/resshift.png} &  
    % % \includegraphics[width=\fwidth]{figs/results/div2k-val/810_color_rect/pasd.png} & 
    % \includegraphics[width=\fwidth]{figs/results/div2k-val/810_color_rect/diffbir.png} & 
    % \includegraphics[width=\fwidth]{figs/results/div2k-val/810_color_rect/seesr.png} &  
    % \includegraphics[width=\fwidth]{figs/results/div2k-val/810_color_rect/SegSR.png} &
    % \includegraphics[width=\fwidth]{figs/results/div2k-val/810_color_rect/HQ.png} 
    % \\ 
     % \centering{\rotatebox{90}{  \small{DIV2K-844}}}  \hspace{1pt} &
    \includegraphics[width=\fwidth]{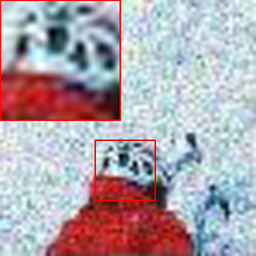} &  
    \includegraphics[width=\fwidth]{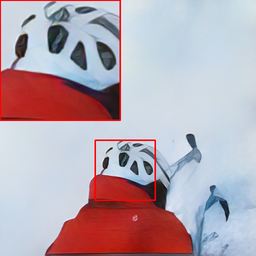} &  
    \includegraphics[width=\fwidth]{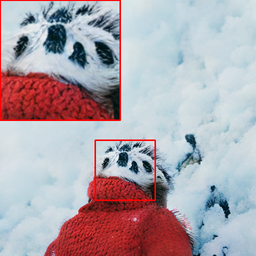} &
    \includegraphics[width=\fwidth]{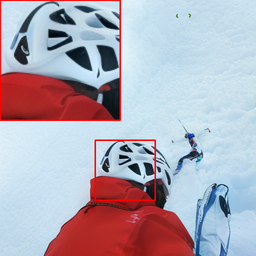} & 
    \includegraphics[width=\fwidth]{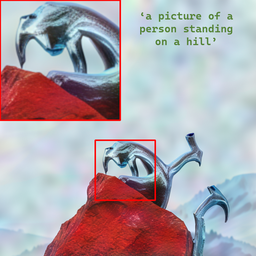} & 
    \includegraphics[width=\fwidth]{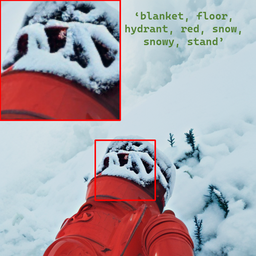} &  
    \includegraphics[width=\fwidth]{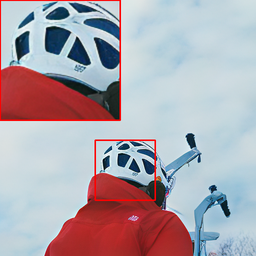} &
    \includegraphics[width=\fwidth]{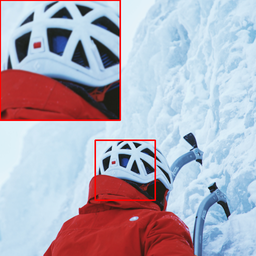} 
    %  \\ 
    %    \small{LQ}
    % % &   \small{BSRGAN}
    % &   \small{Real-ESRGAN}
    % % &   \small{DASR}
    % % &   \small{ResShift}
    % &   \small{StableSR}
    % &  \small{DiffBIR}
    % &   \small{PASD}
    % &   \small{SeeSR}
    % &   \small{\textbf{SegSR}}
    % &   \small{HQ}
    \\ 
    %  \centering{\rotatebox{90}{  \small{OST-071}}}  \hspace{1pt} &
    % \includegraphics[width=\fwidth]{figs/results/ost-val/071_color_rect/LQ.png} &  
    % \includegraphics[width=\fwidth]{figs/results/ost-val/071_color_rect/bsrgan.png} &  
    % \includegraphics[width=\fwidth]{figs/results/ost-val/071_color_rect/realesrgan.png} &  
    % % \includegraphics[width=\fwidth]{figs/results/ost-val/071_color_rect/dasr.png} & 
    % \includegraphics[width=\fwidth]{figs/results/ost-val/071_color_rect/stablesr.png} &  
    % \includegraphics[width=\fwidth]{figs/results/ost-val/071_color_rect/resshift.png} &  
    % % \includegraphics[width=\fwidth]{figs/results/ost-val/071_color_rect/pasd.png} & 
    % \includegraphics[width=\fwidth]{figs/results/ost-val/071_color_rect/diffbir.png} & 
    % \includegraphics[width=\fwidth]{figs/results/ost-val/071_color_rect/seesr.png} &  
    % \includegraphics[width=\fwidth]{figs/results/ost-val/071_color_rect/SegSR.png} &
    % \includegraphics[width=\fwidth]{figs/results/ost-val/071_color_rect/HQ.png} 
     %    
    \includegraphics[width=\fwidth]{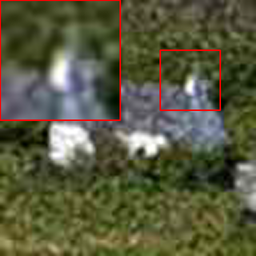} &  
    \includegraphics[width=\fwidth]{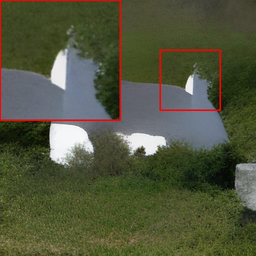} &  
    \includegraphics[width=\fwidth]{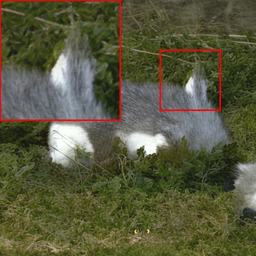} &

    \includegraphics[width=\fwidth]{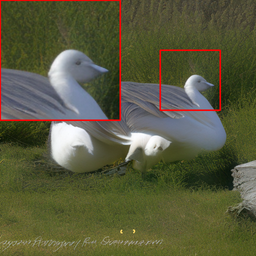} & 
    \includegraphics[width=\fwidth]{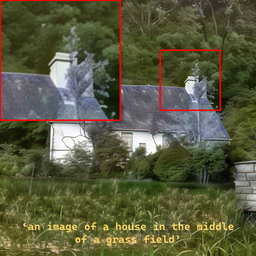} & 
    \includegraphics[width=\fwidth]{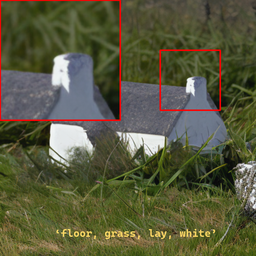} &  
    \includegraphics[width=\fwidth]{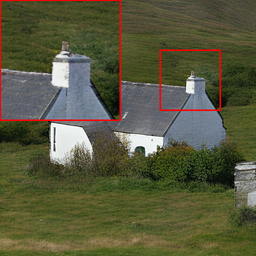} &
    \includegraphics[width=\fwidth]{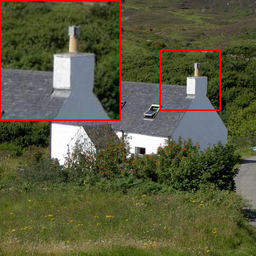} 
     \\ 
       \small{LQ}
    % &   \small{BSRGAN}
    &   \small{Real-ESRGAN}
    % &   \small{DASR}
    % &   \small{ResShift}
    &   \small{StableSR}
    &  \small{DiffBIR}
    &   \small{PASD}
    &   \small{SeeSR}
    &   \small{\textbf{SegSR}}
    &   \small{HQ}
\end{tabular}
\caption{Qualitative comparisons on synthetic benchmarks: DIV2K-Val~\cite{div2k} (top) and OST-Val~\cite{sftgan} (bottom). Please zoom in for details.}
\label{fig:qualitative}
\end{figure*}

\renewcommand{\tabcolsep}{5 pt}

To  achieve  the information flow between the SRDM and SegDM branches so that these two branches can benefit each other during the reverse diffusion process,
we propose a Dual-Modality Bridge (DMB) for joint optimization.
DMB consists of Seg Controller and ImgAided components. 
At time step $t$, the ImgAided component extracts updated image information from $Z_t$ in SRDM, generating a guided feature $I_t$ for Seg Denoiser in SegDM.
Thus, SegDM leverages dynamically updated image information guidance at each step, enabling a more accurate segmentation mask estimation.
Meanwhile, Seg Controller extracts refined segmentation information from $S_t$ in SegDM, and generates semantic control features $C_t$ for SRDM, which is incorporated into the Img Denoiser.
Thus, SRDM leverages refined segmentation condition guidance at each step, facilitating image restoration while preserving semantic structures.
Thanks to the joint optimization strategy facilitated by DMB, the proposed SegSR framework can progressively restore realistic images with accurate semantic details through SRDM, while simultaneously improving the accuracy of semantic segmentation in SegDM.
For more details about the sampling and training strategy of SegSR, please see the supplementary material.

\section{Experiments}
\label{sec:experiments}
\def\fwidth{0.1215\linewidth}
\def\arraystretch{0.5}
\renewcommand{\tabcolsep}{0.5 pt}
\begin{figure*}[!ht]
\centering
\begin{tabular}{ccccccccc}
    % \centering{\rotatebox{90}{  \small{DIV2K-810}}}  \hspace{1pt} &
    % \includegraphics[width=\fwidth]{figs/results/div2k-val/Canon_034_LR4_color_rect_resize/LQ.png} &  
    % \includegraphics[width=\fwidth]{figs/results/div2k-val/Canon_034_LR4_color_rect_resize/bsrgan.png} &  
    % \includegraphics[width=\fwidth]{figs/results/div2k-val/Canon_034_LR4_color_rect_resize/realesrgan.png} &  
    % % \includegraphics[width=\fwidth]{figs/results/div2k-val/Canon_034_LR4_color_rect_resize/dasr.png} & 
    % \includegraphics[width=\fwidth]{figs/results/div2k-val/Canon_034_LR4_color_rect_resize/stablesr.png} &  
    % \includegraphics[width=\fwidth]{figs/results/div2k-val/Canon_034_LR4_color_rect_resize/resshift.png} &  
    % % \includegraphics[width=\fwidth]{figs/results/div2k-val/Canon_034_LR4_color_rect_resize/pasd.png} & 
    % \includegraphics[width=\fwidth]{figs/results/div2k-val/Canon_034_LR4_color_rect_resize/diffbir.png} & 
    % \includegraphics[width=\fwidth]{figs/results/div2k-val/Canon_034_LR4_color_rect_resize/seesr.png} &  
    % \includegraphics[width=\fwidth]{figs/results/div2k-val/Canon_034_LR4_color_rect_resize/SegSR.png} &
    % \includegraphics[width=\fwidth]{figs/results/div2k-val/Canon_034_LR4_color_rect_resize/HQ.png} 
    % \\ 
     % \centering{\rotatebox{90}{  \small{DIV2K-844}}}  \hspace{1pt} &
    \includegraphics[width=\fwidth]{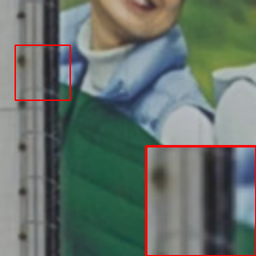} &  
    \includegraphics[width=\fwidth]{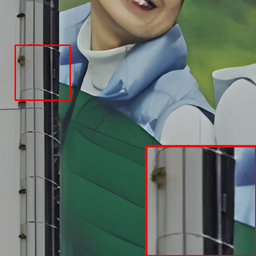} &  
    \includegraphics[width=\fwidth]{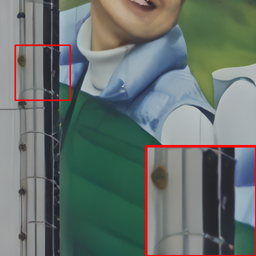} &

    \includegraphics[width=\fwidth]{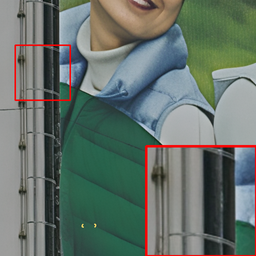}     &
    \includegraphics[width=\fwidth]{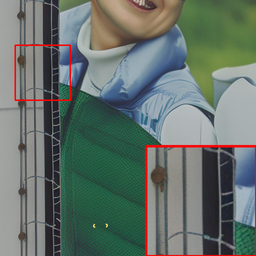} & 
    \includegraphics[width=\fwidth]{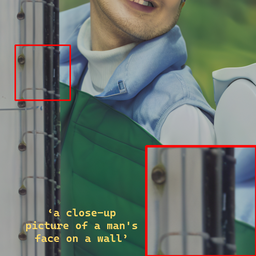} & 
    \includegraphics[width=\fwidth]{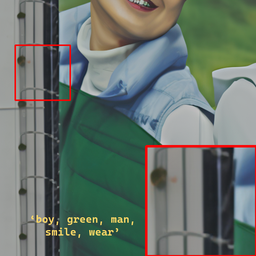} &  
    \includegraphics[width=\fwidth]{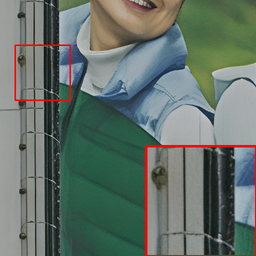} 
    %  \\ 
    %    \small{LQ}
    % % &   \small{BSRGAN}
    % &   \small{Real-ESRGAN}
    % % &   \small{DASR}
    % &   \small{ResShift}
    %  & \small{StableSR}
    %  & \small{DiffBIR}
    % &   \small{PASD}
    % &   \small{SeeSR}
    % &   \small{\textbf{SegSR}}
    \\ 
 
    \includegraphics[width=\fwidth]{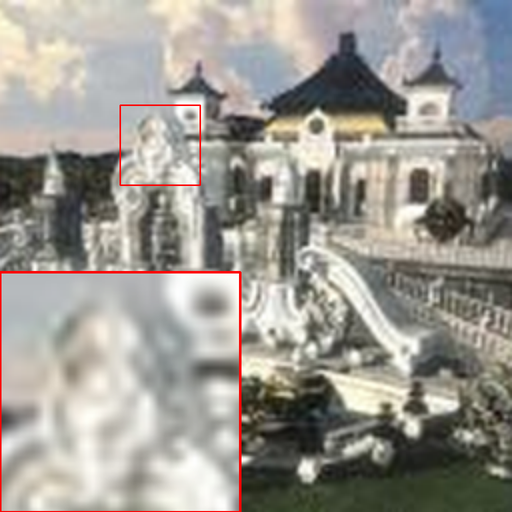} &  
    \includegraphics[width=\fwidth]{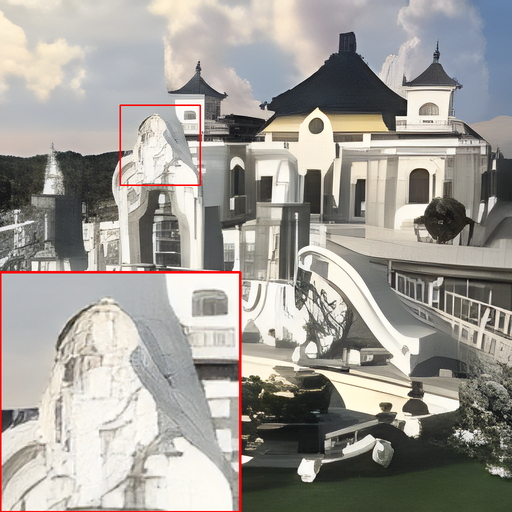} &  
    \includegraphics[width=\fwidth]{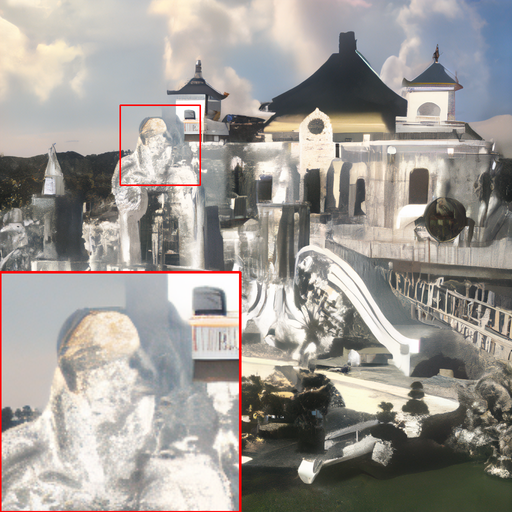}  &
    \includegraphics[width=\fwidth]{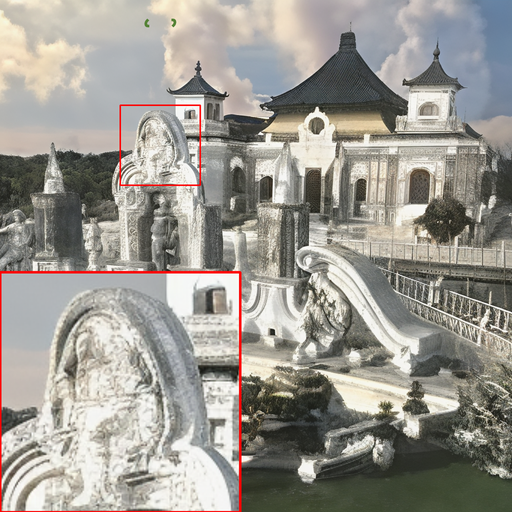} &  
    \includegraphics[width=\fwidth]{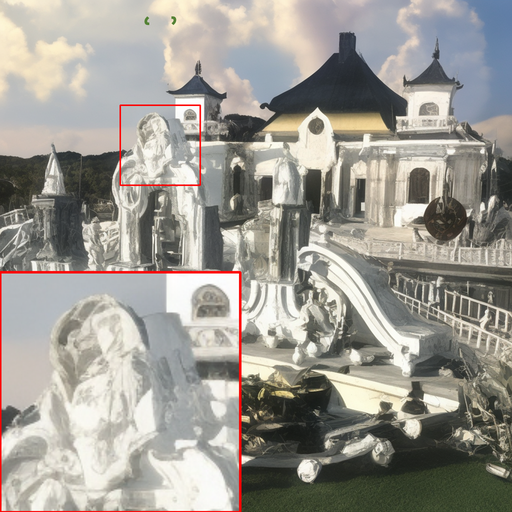} & 
    \includegraphics[width=\fwidth]{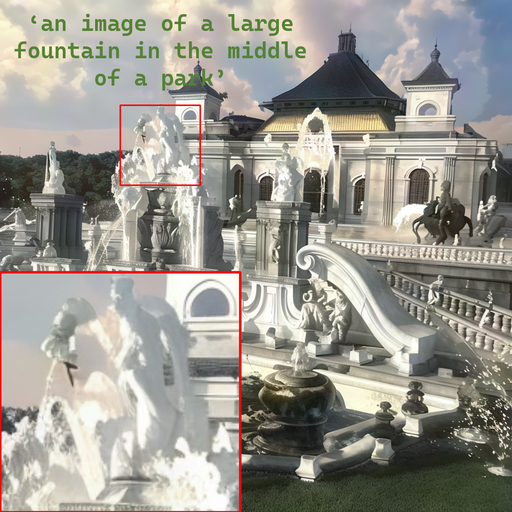} & 
    \includegraphics[width=\fwidth]{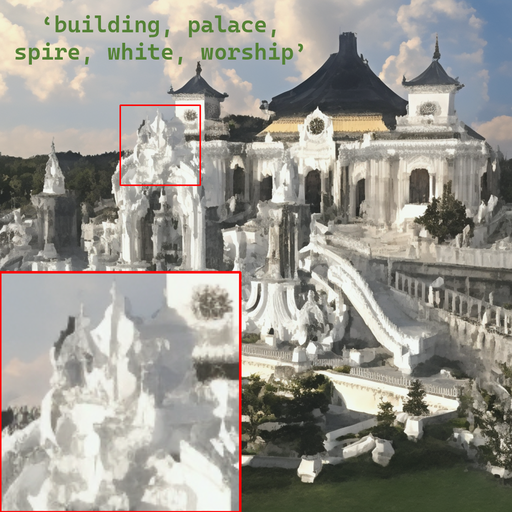} &  
    \includegraphics[width=\fwidth]{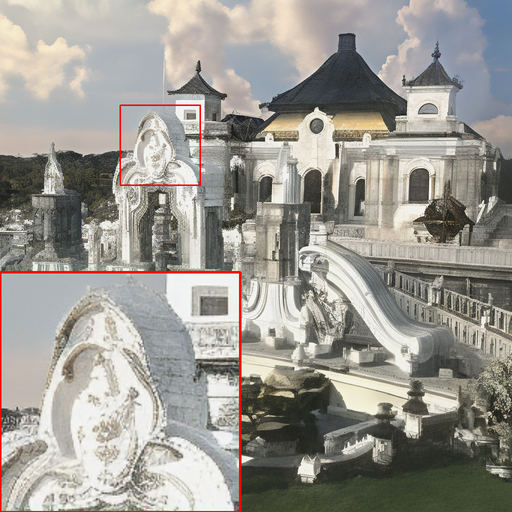} 
    \\

       \small{LQ}
    % &   \small{BSRGAN}
    &   \small{Real-ESRGAN}
    % &   \small{DASR}
    &   \small{ResShift}
      & \small{StableSR}
     & \small{DiffBIR}
    &   \small{PASD}
    &   \small{SeeSR}
    &   \small{\textbf{SegSR}}
    \\  
\end{tabular}
\caption{Qualitative comparisons on real-world benchmarks: RealSR~\cite{realsr} (top)  and RealLQ250~\cite{dreamclear} (bottom).
lease zoom in for details.}
\label{fig:qualitative_real}
\end{figure*}

\renewcommand{\tabcolsep}{5 pt}
\subsection{Experimental Settings}
{\flushleft\textbf{Training Datasets.}}\quad We train our SegSR model using the following datasets: DIV2K~\cite{div2k}, DIV8K~\cite{div8k}, Flickr2K~\cite{frichr2k}, OST~\cite{sftgan}, and  5000 face images from FFHQ~\cite{ffhq}. We adopt the degradation pipeline from Real-ESRGAN~\cite{realesrgan} and the same degradation settings as SeeSR~\cite{seesr} to synthesize LQ-HQ training pairs, For semantic segmentation training, the ground truth segmentation masks are obtained from the pretrained open vocabulary segmentation model, X-Decoder~\cite{x-decoder}. We assume 34 classes representing indoor and outdoor scenes (e.g., sky, mountains, rock, building, tree and etc.), as text prompts for the X-Decoder.
{\flushleft\textbf{Testing Datasets.}}\quad To thoroughly evaluate ISR performance, we conduct testing on both synthetic and real-world datasets. The synthetic datasets are derived from the validation sets of DIV2K~\cite{div2k} and OST~\cite{sftgan}. The images are resized to have the shortest side of 512 pixels, and then center-cropped to 512$\times$512 as the ground truth, followed by applying the same degradation pipeline as SeeSR~\cite{seesr} to generate LQ image. For real-world benchmarks, we use the RealSR~\cite{realsr} and RealLQ250~\cite{drealsr}. The resolution of these two benchmarks is 128$\times$128 and 256$\times$256, respectively. All experiments are conducted on scaling factor $\times4$.

% {\flushleft\textbf{Implementation  Details.}}\quad

% For inference of SegSR, we adopt DDIM~\cite{ddim} with
% 50 sampling steps. The guidance scale  $\omega=5.5.$

{\flushleft\textbf{Evaluation Metrics.}}\quad To comprehensively assess performance, we use both reference and non-reference metrics. PSNR and SSIM~\cite{ssim} (on the Y channel in YCbCr space) measure fidelity, while LPIPS~\cite{lpips} and DISTS~\cite{dists} evaluate perceptual quality. MANIQA~\cite{maniqa}, MUSIQ~\cite{musiq}, and CLIPIQA~\cite{clipiqa} serve as non-reference quality metrics.

\subsection{Comparisons with State-of-the-Art Methods}

{\flushleft\textbf{Compared Methods.}}\quad We conduct a comparative analysis of our SegSR with other state-of-the-art (SOTA) Real-ISR methods, including GAN-based methods BSRGAN~\cite{bsrgan}, Real-ESRGAN~\cite{realesrgan} and DASR~\cite{dasr}, as well as diffusion-based methods like StableSR~\cite{stablesr}, ResShift~\cite{resshift}, DiffBIR~\cite{diffbir}, PASD~\cite{pasd}, and SeeSR~\cite{seesr}. For testing, we employ the publicly available implementations and pretrained models of these competing methods.

{\flushleft\textbf{Quantitative Comparisons.}}
Table~\ref{tab:quantitative} presents the quantitative results on various synthetic and real-world benchmarks. Our method demonstrates strong performance across no-reference metrics (MANIQA, MUSIQ and CLIPIQA),  underscoring the high quality of our restorations. 
As observed, GAN-based methods consistently excel in PSNR/SSIM scores, offering higher fidelity for LQ images but often lack realistic detail generation. However, as previous studies have highlighted~\cite{supir,xpsr,dreamclear}, full-reference metrics may not accurately capture human preferences. In contrast, diffusion-based methods focus on photorealistic restoration, typically yet they lag in low scores in full-reference metrics like PSNR/SSIM, possibly due to their strong generative capacity for realistic details not present in ground truth images.

\begin{table*}[t]
    \centering
    \resizebox{\textwidth}{!}{%
    \begin{NiceTabular}{c|c|c|c|c|c|c|c|c|c|c}
    \toprule
    \multirow{2}{*}{\textbf{Method}} & \multicolumn{5}{c|}{\textbf{Settings}} & \multicolumn{6}{c}{\textbf{DIV2K-Val / RealSR \cite{realsr}}} \\
    % \midrule
    & \textbf{SRDM} & \textbf{Segformer} & \textbf{SegDM} & \textbf{DMB} &\textbf{HQ-Seg} & \textbf{PSNR $\uparrow$}  & \textbf{MUSIQ $\uparrow$} & \textbf{MANIQA $\uparrow$} & \textbf{CLIPIQA $\uparrow$} & \textbf{ACC $\uparrow$}\\
    \midrule
    Exp.(1)  & \checkmark & \texttimes & \texttimes & \texttimes  & \texttimes & 20.38 / 24.65  & 71.35 / 66.16 & 0.5341 / 0.4506 & 0.7519 / 0.6356 & 0.6621 / 0.5753\\ 
    Exp.(2)  & \checkmark & \checkmark & \texttimes & \texttimes  & \texttimes & 20.35 / 24.69  & 72.34 / 67.22 & 0.5812 / 0.4898 & 0.7699 / 0.6692 &  0.6703 / 0.6227 \\
    Exp.(3)  & \checkmark & \checkmark & \checkmark & \texttimes  & \texttimes & 20.37 / 24.64 & 72.40 / 66.83 & 0.5840 / 0.4968 & 0.7672 / 0.6845  &  0.7009 /  0.6357\\
     Exp.(4)  & \checkmark & \checkmark & \checkmark & \checkmark & \texttimes & 20.42 / 24.61 & 72.29 / 67.80 & 0.6006 / 0.5233 & 0.7723 / 0.6906  &0.7116 / 0.6396  \\
     Exp.(5)  & \checkmark & \texttimes & \texttimes & \texttimes  & \checkmark &20.39 / 24.83 & 72.38 / 67.95 &  0.6040 / 0.5241 &0.7777 / 0.6958  &0.8039 / 0.6472 \\
    \bottomrule
    \end{NiceTabular}}
    \caption{Ablation study to validate the effectiveness of different components. For the detailed experimental settings, please refer to Sec~\ref{sec:ablation}.}
    \label{tab:ablation}
\end{table*}

% \begin{table*}[t]
%     \centering
%     \label{tab:ablation}
%     \resizebox{\textwidth}{!}{%
%     \begin{NiceTabular}{c|l|c|c|c|c|c}
%     \toprule
%     \multirow{2}{*}{\textbf{Method}} & \multirow{2}{*}{\textbf{Settings}} & \multicolumn{6}{c}{\textbf{DIV2K-Val / RealSR \cite{realsr}}} \\
%     % \midrule
%     && \textbf{PSNR $\uparrow$}  & \textbf{MUSIQ $\uparrow$} & \textbf{MANIQA $\uparrow$} & \textbf{CLIPIQA $\uparrow$} & \textbf{ACC $\uparrow$}\\
%     \midrule
%     Exp.(1) &SRDM & 20.38 / 24.65  & 71.35 / 66.16 & 0.5341 / 0.4506 & 0.7519 / 0.6356 & 0.6621 / 5.8707\\
%      Exp.(2) &SRDM + Segformer   & 20.35 / 24.69  & 72.34 / 67.22 & 0.5812 / 0.4898 & 0.7699 / 0.6692 &  0.6703 / 5.6319 \\
%     Exp.(3)&SRDM + SegDM  & 20.37 / 24.64 & 72.40 / 66.83 & 0.5840 / 0.4968 & 0.7672 / 0.6845  &  0.7009 / 5.6319\\
%      Exp.(4)&SRDM +    & 20.42 / 24.61 & 72.29 / 67.80 & 0.6006 / 0.5233 & 0.7723 / 0.6906  &0.7116 / 5.5182  \\
%      Exp.(5)&SRDM   &20.39 / 24.61 & 72.38 / 67.80 &  0.6040 / 0.5233 &0.7777 / 0.6906  &0.8039 / 5.5182 \\
%     \bottomrule
%     \end{NiceTabular}}
%     \caption{Results of different methods compared under different settings.}
% \end{table*}
\begin{figure*}[!t]
    \centering
    \includegraphics[width=\linewidth]{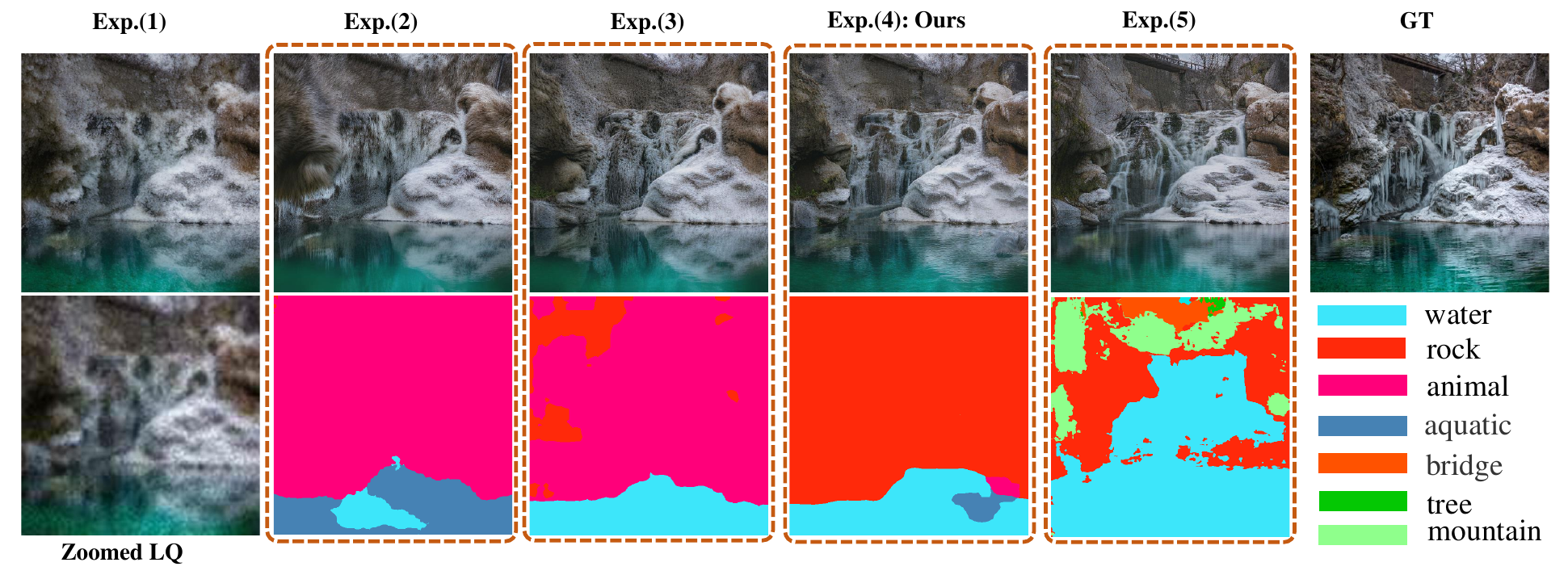}
    \caption{Ablation study to validate the effectiveness of initial Segformer, SegDM, and DMB. The segmentation masks shown below each image super-resolution result represent the segmentation predictions used for the image restoration task. In regions where the segmentation masks are inaccurately estimated, the image restoration process produces incorrect semantic details. For the detailed settings of different methods, please refer to Sec.~\ref{sec:ablation}.}
    \label{fig:ablation}
\end{figure*}

{\flushleft\textbf{Qualitative Comparisons.}}
Figures~\ref{fig:qualitative} and~\ref{fig:qualitative_real} show visual comparisons of synthetic and real-world images, respectively. As shown in Figure~\ref{fig:qualitative}, when the image suffers from severe degradation, the results of Real-ESRGAN tend to be over-smooth and lack details.  Conditioned on LQ images without additional semantic control, StableSR generates results with obvious semantic errors, such as animal fur on the house region (row 1). 
Due to the initial restoration for LQ image, DiffBIR produces more realistic results than StableSR (row 1) but still introduces semantic errors, such as turning a house into a swan (row 2). 
PASD's lack of clear positional information causes a `hill' style to appear in the `person' region (row 1). 
Similarly, SeeSR fails to recognize the person component and mistakenly considers them as `hydrant', leading to hydrant-like textures in the person region (row 1).
In comparison, semantic segmentation enables a more comprehensive perception of salient objects within an image and facilitates clearer semantic spatial localization, aiding SegSR to generate realistic images and accurate semantic details. 

When it comes to real-world images in Figure~\ref{fig:qualitative_real}, Real-ESRGAN still struggles with generating realistic details. StableSR and DiffBIR exhibit artifacts in smooth regions (row 1), while StableSR shows visually unpleasant over-generation (row 2). Other methods, such as PASD and SeeSR, may produce blurry results. These methods fail to recognize the building component, resulting in the absence of corresponding semantic texture details in the zoomed-in region (row 1). Additionally, they may generate distorted structures due to the lack of clear positional information, leading to incorrect semantic structure generation in the zoomed-in region (row 2). In contrast, SegSR produces sharper and more accurate semantic details, such as the edges of the building (row 2). More visual examples can be found in the supplementary material.

\subsection{Ablation Studies}
\label{sec:ablation}

In this subsection, we validate the effectiveness of each component in the proposed method, with results presented in Table~\ref{tab:ablation} and Figure~\ref{fig:ablation}.
Exp. (1) includes only SRDM conditioned on the LQ image. 
As seen in the first column of the top row in Figure~\ref{fig:ablation}, the restored results exhibit incorrect, fluff-like artifacts due to the absence of semantic control during the diffusion process.
In Exp. (2), we introduce the initial segmentation predictions from Segformer as guidance to SRDM.
As shown in the second column of Figure~\ref{fig:ablation}, the segmentation mask estimated by Segformer remains inaccurate, even after finetuning on LQ images.
Consequently, the restoration results, guided by this inaccurate mask, contain unrealistic details, such as "animal-like" textures appearing in the rock region.
Exp. (3) uses SegDM, which is conditioned on initial image representation predictions from Segformer, to predict segmentation conditions for SRDM.
Thanks to the strong segmentation distribution modeling ability from SegDM, Exp. (3) achieves improved semantic segmentation over Exp. (2), as shown in the third column of Figure~\ref{fig:ablation}.
However, a significant portion of the segmentation mask in the rock areas remains inaccurate.
This is because SegDM in Exp. (3) does not leverage the higher-quality image information from SRDM to refine semantic segmentation predictions during the diffusion process.
Our method introduces the DMB module, enabling SegDM to utilize high-quality images from SRDM to correct inaccurately estimated semantic segmentation.
Meanwhile, SRDM can restore images with better semantic details by leveraging the updated semantic segmentation from SegDM as shown in the fourth column of Figure~\ref{fig:ablation}. 
In addition, we also analyze the performance upper bound brought by the semantic segmentation prior. 
In  Exp. (5), we use X-decoder~\cite{x-decoder} to perform segmentation prediction on the HQ images.
As shown in the fifth column of Figure~\ref{fig:ablation}, the segmentation predictions for the HQ images help SRDM generate realistic images with correct semantic details.
Similarly, Table~\ref{tab:ablation} shows that the proposed method can achieve consistently better performance with more components considered which demonstrates the effectiveness of Segformer, SegDM, and DMB. 
The ACC metric represents the accuracy of semantic segmentation predictions by X-Decoder on the restoration results from SRDM, using the predictions on HQ images as the ground truth mask. 
Notably, in both quantitative and qualitative results for semantic segmentation prediction and image restoration, our method closely approximates the performance of Exp. (5).

\section{Conclusion}
\label{sec:conclusion}
In this paper, we propose SegSR, a diffusion-based approach that introduces semantic segmentation as an additional control condition for real-world image super-resolution. In practice, we develop a dual-diffusion framework with a Dual-Modality Bridge module to facilitate interaction between the image super-resolution and segmentation diffusion models. By leveraging semantic segmentation, SegSR enables more comprehensive object perception and facilitates clearer semantic spatial localization.  Extensive experiments show that SegSR generates realistic images while effectively preserving semantic structures.

% We introduce SegSR, an approach that leverages semantic segmentation prior to that enables comprehensive salient object perception and explicitly allocate objects to their respective spatial regions.
% As improved image results and segmentation
% predictions can mutually enhance each other’s quality, we introduces a dual-diffusion
% framework to facilitate interaction between the Real-ISR and segmentation diffusion models for optimal cooperative restoration.

% we trained a robust, discrete diffusion-based segmentor capable of producing reliable semantic priors to guide the image super-resolution process. Furthermore, recognizing the mutual influence between image quality and segmentation accuracy, we designed a CMA module that facilitates their joint optimization, leading to superior restoration results. Our method marks a significant step towards more effective utilization of generative priors for producing semantically and spatially accurate Real-ISR images, as demonstrated by extensive experimental validation.
{
    \small
    \bibliographystyle{ieeenat_fullname}
    \bibliography{main}
}

% WARNING: do not forget to delete the supplementary pages from your submission 
% \input{sec/X_suppl}

\end{document}